\documentclass{article}
\pdfoutput=1
\usepackage{times}
\usepackage{graphicx} 
\usepackage{subcaption}
\usepackage{amsmath}
\usepackage{amssymb}

\usepackage{natbib}

\usepackage{algorithm}
\usepackage{algorithmic}

\usepackage{hyperref}

\newcommand{\pd}[2]{\frac{\partial #1}{\partial #2}}

\newcommand{\mb}{\mathbf}

\newcommand{\mc}{\mathcal}

\newcommand{\expect}[2]{\mathbb{E}_{#1}\left[ #2 \right]}

\usepackage{xcolor}
\definecolor{darkgreen}{rgb}{0,0.6,0}

\usepackage[accepted]{icml2017}

\icmltitlerunning{Learned Optimizers that Scale and Generalize}

\begin{document} 
\twocolumn[
\icmltitle{
Learned Optimizers that Scale and Generalize
}



\icmlsetsymbol{i}{*}

\begin{icmlauthorlist}
\icmlauthor{Olga Wichrowska}{brain}
\icmlauthor{Niru Maheswaranathan}{int,stanford}
\icmlauthor{Matthew W. Hoffman}{dm}
\icmlauthor{Sergio G\'omez Colmenarejo}{dm}
\icmlauthor{Misha Denil}{dm}
\icmlauthor{Nando de Freitas}{dm} 
\icmlauthor{Jascha Sohl-Dickstein}{brain}
\end{icmlauthorlist}

\icmlaffiliation{brain}{Google Brain}
\icmlaffiliation{stanford}{Stanford University}
\icmlaffiliation{dm}{Deepmind}
\icmlaffiliation{int}{Work done during an internship at Google Brain.}

\icmlcorrespondingauthor{Olga Wichrowska}{olganw@google.com}

\icmlkeywords{learning to learn, machine learning, optimization}

\vskip 0.3in
]



\printAffiliationsAndNotice{}

\begin{abstract} 
Learning to learn has emerged as an important direction for achieving artificial intelligence. Two of the primary barriers to its adoption are an inability to scale to larger problems and a limited ability to generalize to new tasks. We introduce a learned gradient descent optimizer that generalizes well to new tasks, and which has significantly reduced memory and computation overhead. We achieve this by introducing a novel hierarchical RNN architecture, with minimal per-parameter overhead, augmented with additional architectural features that mirror the known structure of optimization tasks. We also develop a meta-training ensemble of small, diverse optimization tasks capturing common properties of loss landscapes. The optimizer learns to outperform RMSProp/ADAM on problems in this corpus. More importantly, it performs comparably or better when applied to small convolutional neural networks, despite seeing no neural networks in its meta-training set. Finally, it generalizes to train Inception V3 and ResNet V2 architectures on the ImageNet dataset for thousands of steps, optimization problems that are of a vastly different scale than those it was trained on. We release an open source implementation of the meta-training algorithm.
\end{abstract} 

\section{Introduction}

Optimization is a bottleneck for almost all tasks in machine learning, as well as in many other fields, including engineering, design, operations research, and statistics. Advances in optimization therefore have broad impact.
Historically, optimization has been performed using hand-designed algorithms.
Recent results in machine learning show that, given sufficient data, well-trained neural networks often outperform hand-tuned approaches on supervised tasks. This raises the tantalizing possibility that neural networks may be able to outperform hand-designed optimizers.

Despite the promise in this approach, previous work on learned RNN optimizers for gradient descent has failed to produce neural network optimizers that generalize to new problems, or that continue to make progress on the problems for which they were meta-trained when run for large numbers of steps (see Figure \ref{fig opt diverge}). Current neural network optimizers are additionally too costly in both memory and computation to scale to larger problems.

We address both of these issues. Specifically, we improve upon existing learned optimizers by:
\begin{enumerate}
    \item Developing a meta-training set that consists of an ensemble of small tasks with diverse loss landscapes
    \item Introducing a hierarchical RNN architecture with lower memory and compute overhead, and which is capable of capturing inter-parameter dependencies.
    \item Incorporating features motivated by successful hand-designed optimizers into the RNN, so that it can build on existing techniques. These include dynamically adapted input and output scaling, momentum at multiple time scales, and a cross between Nesterov momentum and RNN attention mechanisms.
    \item Improving the meta-optimization pipeline, for instance by introducing a meta-objective that better encourages exact convergence of the optimizer, and by drawing the number of optimization steps during training from a heavy tailed distribution. 
\end{enumerate}

\section{Related work}



Learning to learn has a long history in psychology \citep{ward1937reminiscence,harlow1949formation,kehoe1988layered,lake:2016}. 
Inspired by it, machine learning researchers have proposed meta-learning techniques for 
optimizing the process of learning itself. 
\citet{schmidhuber:1987}, for example, considers networks that are able to modify their own weights.  This leads to end-to-end differentiable systems which allow, in principle, for extremely general update strategies to be learned. There are many works related to this idea, including
\citep{sutton:1992,naik:1992,thrun:1998,hochreiter:2001,santoro:2016}.  

A series of papers from \citet{bengio:1990,Bengio+al-92,bengio:1995}
presents methods for learning parameterized local neural network update rules that avoid back-propagation. \citet{runarsson:2000} extend this to more complex update models.  The result of meta learning in these cases is an algorithm, i.e.\ a local update rule.

\citet{andrychowicz2016learning} learn to learn by gradient descent by gradient descent.  Rather than trying to distill a global objective into a local rule, their work focuses on learning how to integrate gradient observations over time in order to achieve fast learning of the model.  The component-wise structure of the algorithm allows a single learned algorithm to be applied to new problems of different dimensionality.
While  \citet{andrychowicz2016learning} consider the issue of transfer to different datasets and model structures, they focus on transferring to problems of the same class. In fact, they report negative results when transferring optimizers, meta-trained to optimize neural networks with logistic functions, to networks with ReLU functions.

\citet{Li2017learning} proposed an approach similar to \citet{andrychowicz2016learning}, around the same time, but they rely on policy search to compute the meta-parameters of the optimizer. That is, they learn to learn by gradient descent by reinforcement learning.

\citet{ZophLe2017} also meta-train a controller RNN, but this time to produce a string in a custom domain specific language (DSL) for describing neural network architectures.  An architecture matching the produced configuration (the ``child'' network) is instantiated and trained in the ordinary way.  In this case the meta-learning happens only at the network architecture level.  

\citet{Ravi2017optimization} modify the optimizer of \citet{andrychowicz2016learning} for 1 and 5-shot learning tasks. They use test error to optimize the meta learner. These tasks have the nice property that the recurrent neural networks only need to be unrolled for a small number of steps.

\citet{Wang2016learning} show that it is possible to learn to solve reinforcement learning tasks by reinforcement learning. They demonstrate their approach on several examples from the bandits and cognitive science literature. A related approach was proposed by \citet{Duan2016}.

Finally, \citet{Chen2016learning} also learn reinforcement learning, but by supervised meta-training of the meta-learner. They apply their methods to black-box function optimization tasks, such as Gaussian process bandits, simple low-dimensional controllers, and hyper-parameter tuning.

\section{Architecture}\label{sec arch}

\begin{figure}[h]
\includegraphics[width=\columnwidth]{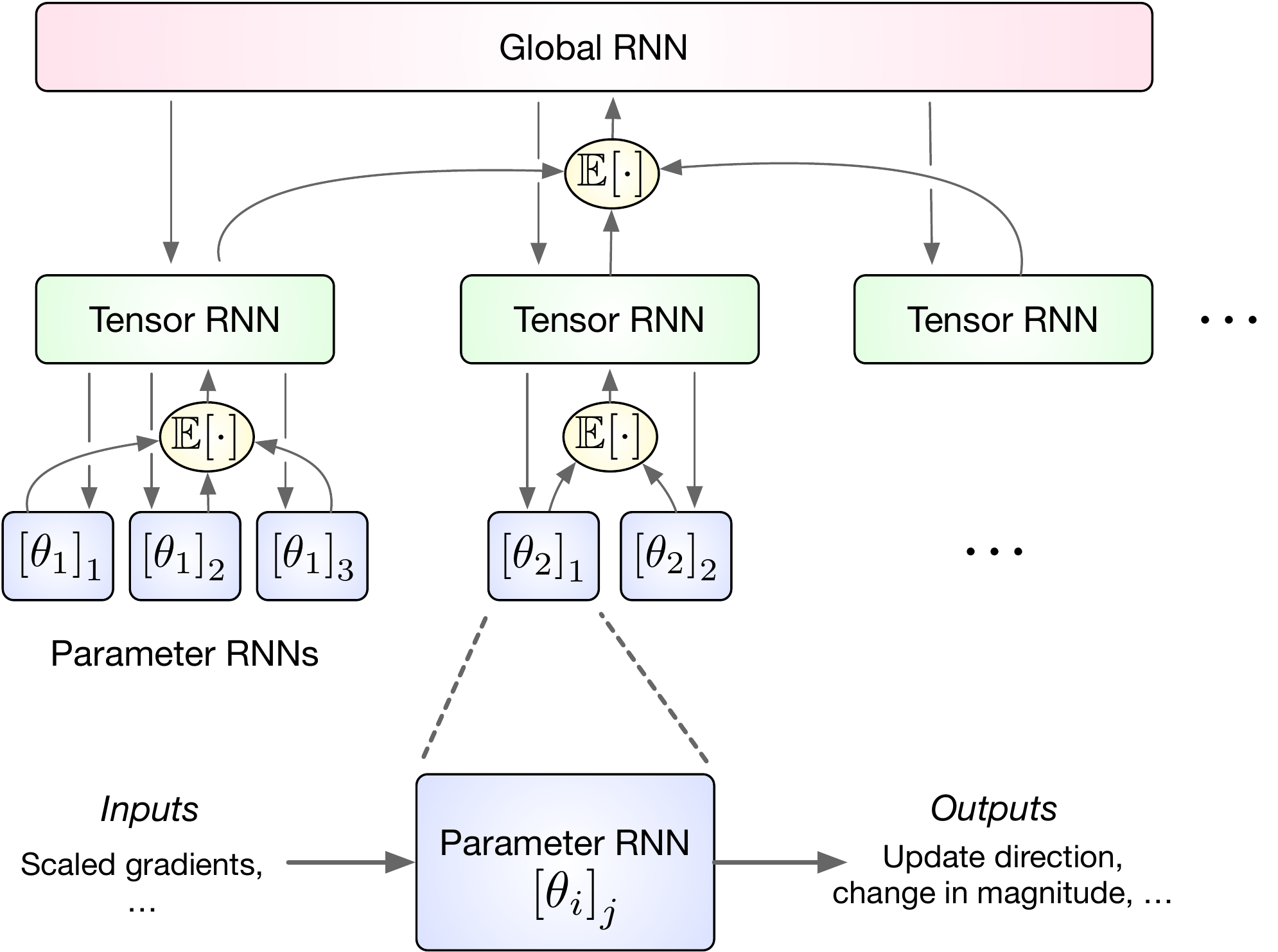}
\caption{
Hierarchical RNN architecture. 
At the lowest level, a small Parameter RNN processes the inputs and outputs (Section \ref{sec in out}) for every parameter ($\theta_{ij}$) in the target problem. 
At the intermediate level, a medium-sized Tensor RNN exists for every parameter tensor (denoted by $\theta_i$) in the target problem. It takes as input the average latent state across all Parameter RNNs belonging to the same tensor. Its output enters those same Parameter RNNs as a bias term.
At the top level, a single Global RNN receives as input the average hidden state of all Parameter RNNs, and its output enters the Tensor RNNs as a bias term and is added to the Parameter RNN bias term. This architecture has low per-parameter overhead, while the Tensor RNNs are able to capture inter-parameter dependencies, and the Global RNN is able to capture inter-tensor dependencies.}
\label{fig arch diagram}
\end{figure}

At a high level, a hierarchical RNN is constructed to act as a learned optimizer, 
with its architecture matched to the parameters in the target problem. 
The hierarchical RNN's parameters (called meta-parameters) are shared across all target problems, so despite having an architecture that adapts to the target problem, it can be applied to new problems.
At each optimization step, the learned optimizer receives the gradients for every parameter along with some additional quantities derived from the gradients, and outputs an update to the parameters. Figure \ref{fig arch diagram} gives an overview. 


\subsection{Hierarchical architecture}
\label{sec hier}
In order to effectively scale to large problems, the optimizer RNN must stay quite small while maintaining enough flexibility to capture inter-parameter dependencies that shape the geometry of the loss surface.
Optimizers that account for this second order information are often particularly effective (e.g. quasi-Newton approaches).
We propose a novel hierarchical architecture to enable both low per-parameter computational cost, and aggregation of gradient information and coordination of update steps across parameters (Figure \ref{fig arch diagram}). 
At the lowest level of the hierarchy, we have a small \textit{Parameter RNN} that receives direct per-parameter (scalar) gradient inputs. One level up, we have an intermediate \textit{Tensor RNN} that incorporates information from a subset of the Parameter RNNs (where the subsets are problem specific). For example, consider a feedforward fully-connected neural network. There would be a Tensor RNN for each layer of the network, where each layer contains an ($n \times m$) weight matrix and therefore $nm$ Parameter RNNs.

At the highest level of the hierarchy is a \textit{Global RNN} which receives output from every Tensor RNN. This allows the Parameter RNN to have very few hidden units with larger Tensor and Global RNNs keeping track of problem-level information. The Tensor and Global RNNs can also serve as communication channels between Parameter and Tensor RNNs respectively. The Tensor RNN outputs are fed as biases to the Parameter RNN, and the new parameter state is averaged and fed as input to the Tensor RNN. Similarly, the Global RNN state is fed as a bias to each Tensor RNN, and the output of the Tensor RNNs is averaged and fed as input to the Global RNN (Figure \ref{fig arch diagram}).

The architecture used in the experimental results has a Parameter RNN hidden state size of 10, and a Tensor and Global RNN state size of 20 (the architecture used by \citet{andrychowicz2016learning} had a two layer RNN for each parameter, with 20 units per layer). These sizes showed the best generalization to ConvNets and other complex test problems. Experimentally, we found that we could make the Parameter RNN as small as 5, and the Tensor RNN as small as 10 and still see good performance on most problems. We also found that the performance decreased slightly even on simple test problems if we removed the Global RNN entirely. 
We used a GRU architecture
\citep{cho2014properties} 
for all three of the RNN levels.

\subsection{Features inspired by optimization literature}\label{sec features}


The best performing neural networks often have knowledge about task structure baked into their design. Examples of this include convolutional models for image processing \citep{krizhevsky2012imagenet, he2016identity}, causal models (RNNs) for modeling causal time series data, and the merging of neural value functions with Monte Carlo tree search in AlphaGo \citep{silver2016mastering}.




We similarly incorporate knowledge of effective strategies for optimization into our network architecture. 
We emphasize that these are not arbitrary design choices. 
The features below are motivated by results in optimization and recurrent network literature.
They are also individually important to the ability of the learned optimizer to generalize to new problems, as is
illustrated by the ablation study in Section \ref{sec ablation} and Figure \ref{fig ablation}.


Let $L\left(\theta \right)$ be the loss of the target problem, where $\theta = \left\{ \theta_1, ..., \theta_{N_T} \right\}$ is the set of all parameter tensors $\theta_t$ (e.g. all weight matrices and bias vectors in a neural network). 
At each training iteration $n$, each parameter tensor $t$ is updated as $\theta^{n+1}_t = \theta^n_t + \Delta \theta_t^n$, where the update step $\Delta \theta_t^n$ is set by the learned optimizer (Equation \ref{eq update step} below).

\subsubsection{Attention and Nesterov Momentum} \label{sec attention}

Nesterov momentum \citep{nesterov1983method} is a powerful optimization approach, where parameter updates are based not on the gradient evaluated at the current iterate $\theta^n$, but rather at a location $\phi^n$ which is extrapolated ahead of the current iterate. 
Similarly, attention mechanisms have proven extremely powerful in recurrent translation models \citep{bahdanau2015neural}, decoupling the iteration $n$ of RNN dynamics from the observed portion of the input sequence. 
Motivated by these successes, we incorporate an attention mechanism that allows the optimizer to 
explore new regions of the loss surface by computing gradients away (or ahead) from the current parameter position.
At each training step $n$ the attended location is set as $\phi^{n+1}_t = \theta^n_t + \Delta \phi_t^n$, where the offset $\Delta \phi_t^n$ is further described by Equation \ref{eq attend step} below. Note that the attended location is an offset from the previous parameter location $\theta^n$ rather than the previous attended location $\phi^n$. 

The gradient $\mb g^n$ of the loss $L\left(\theta \right)$ with respect to the attended parameter values $\phi^n$ will provide the only input to the learned optimizer, though it will be further transformed before being passed to the hierarchical RNN. For every parameter tensor $t$, $\mb g^n_t = \pd{L}{\phi^n_t}$.





\subsubsection{Momentum on multiple timescales} \label{sec multi_timescale}

Momentum with an exponential moving average is typically motivated in terms of averaging away minibatch noise or high frequency oscillations, and is often a very effective feature \citep{nesterov:1983,tseng:1998}. 
We provide the learned optimizer with exponential moving averages $\bar{\mb g}_{ts}$ of the gradients on several timescales, where $s$ indexes the timescale of the average. The update equation for the moving average is
\begin{align}
    \bar{\mb g}_{ts}^{n+1} &=
        \bar{\mb g}_{ts}^n \sigma\left(\beta_{gt}^n\right)^{2^{-s}}  + 
        g_t^n \left(1 - \sigma\left(\beta_{gt}^n\right)^{2^{-s}} \right)
,
\end{align}
where the $\sigma$ indicates the sigmoid function, and where the momentum logit $\beta^n_{g t}$ for the shortest $s=0$ timescale is output by the RNN, and the remaining timescales each increase by a factor of two from that baseline.

By comparing the moving averages at multiple timescales, the learned optimizer has access to information about how rapidly the gradient is changing with training time (a measure of loss surface curvature), and about the degree of noise in the gradient.

\subsubsection{Dynamic input scaling}\label{sec dyn inp scl}

We would like our optimizer to be invariant to parameter scale. 
Additionally, RNNs are most easily trained when their inputs are well conditioned, and have a similar scale as their latent state. 
In order to aid each of these goals, we rescale the average gradients in a fashion similar to what is done in RMSProp \citep{tieleman:2012}, ADAM \citep{kingma2015adam}, and SMORMS3 \citep{smorms3},
\begin{align}
    \lambda^{n+1}_{ts} &= 
        \lambda^{n}_{ts}
            \sigma\left(\beta_{\lambda t}^n\right)^{2^{-s}}  + 
        \left(\bar{\mb g}_{ts}^n\right)^2 \left(1 - \sigma\left(\beta_{\lambda t}^n\right)^{2^{-s}} \right) \\
    \mb m^n_{ts} &= \frac{ \bar{\mb g}_{ts}^n }{ \sqrt{\lambda^{n}_{ts}} } \label{eq grads scaled}
,
\end{align}
where $\lambda^{n}_{ts}$ is a running average of the square {\em average} gradient, $\mb m^n_{ts}$ is the scaled averaged gradient, and the momentum logit $\beta_{\lambda t}^n$ for the shortest $s=0$ timescale will be output by the RNN, similar to how the timescales for momentum are computed in the previous section.

It may be useful for the learned optimizer to have access to how gradient magnitudes are changing with training time. We therefore provide as further input a measure of {\em relative} gradient magnitudes at each averaging scale $s$. Specifically, we provide the relative log gradient magnitudes,
\begin{align}
\gamma^n_{ts} &= \log \lambda^{n}_{ts} - \expect{s}{\log \lambda^{n}_{ts}}.
\end{align}

\subsubsection{Decomposition of output into direction and step length}\label{sec decomposition}

Another aspect of RMSProp and ADAM is that the learning rate corresponds directly to the characteristic step length. This is true because the gradient is scaled by a running estimate of its standard deviation, and after scaling has a characteristic magnitude of 1. The length of update steps therefore scales linearly with the learning rate, but is invariant to any scaling of the gradients.

We enforce a similar decomposition of the parameter updates into update directions $\mb d^n_\theta$ and $\mb d^n_\phi$ for parameters and attended parameters, with corresponding step lengths $\exp\left(\eta^n_\theta\right)$ and $\exp\left(\eta^n_\phi\right)$,
\begin{align}
    \Delta \mb \theta^n_t &= \exp\left(\eta^n_{\theta t}\right) \frac{ 
        \mb d^n_{\theta t}
        }{
        \left|\left| \mb d^n_{\theta t} \right|\right| / N_t
        }
        \label{eq update step} 
        , \\
    \Delta \mb \phi^n_t &= \exp\left(\eta^n_\phi\right) \frac{ 
        \mb d^n_{\phi t}
        }{
        \left|\left| \mb d^n_{\phi t} \right|\right| / N_t
        }
        \label{eq attend step}
        ,
\end{align}
where $N_t$ is the number of elements in the parameter tensor $\theta_t$. 
The directions $\mb d^n_{\theta t}$ and $\mb d^n_{\phi t}$ are read directly out of the RNN (though see \ref{sec shortcut} for subtleties).

\paragraph{Relative learning rate}
We want the performance of the optimizer to be invariant to parameter scale. This requires that the optimizer judge the correct step length from the history of gradients, rather than memorizing the range of step lengths that were useful in its meta-training ensemble. 
The RNN therefore controls step length by outputing a multiplicative (additive after taking a logarithm) change, rather than by outputing the step length directly,
\begin{align}
    \eta^{n+1}_\theta &= \Delta \eta^n_\theta + \bar{\eta}^{n+1}_\theta, \\
\bar{\eta}^{n+1}_\theta &= \gamma \bar{\eta}^n_\theta + \left(1 - \gamma\right) \eta^{n+1}_\theta   ,
\end{align}
where for stability reasons, the log step length $\eta^n_\theta$ is specified relative to an exponential running average $\bar{\eta}^n_\theta$ with meta-learned momentum $\gamma$. 
The attended parameter log step length $\eta^n_\theta$ is related to $\eta^n_\theta$ by a meta-learned constant offset $c$,
\begin{align}
    \eta^{n}_\phi &= \eta^n_\theta + c
    .
\end{align}


To further force the optimizer to dynamically adapt the learning rate rather than memorizing a learning rate trajectory, the learning rate is initialized from a log uniform distribution from $10^{-6}$ to $10^{-2}$. We emphasize that the RNN has no direct access to the learning rate, so it must adjust it based purely on its observations of the statistics of the gradients.

In order to aid in coordination across parameters, we do provide the RNN as an input the {\em relative} log learning rate of each parameter, compared to the remaining parameters, $\eta^n_\text{rel} = \eta^n_\theta - \expect{ti}{\eta^n_{\theta t i}}$.







\subsection{Optimizer inputs and outputs}
\label{sec in out}
As described in the preceding sections, the full set of Parameter RNN inputs for each tensor $t$ are 
$\mb x^n_t = \left\{\mb m^n_{t}, \mb \gamma^n_t,  \mb \eta^n_\text{rel} \right\}$,
corresponding to the scaled averaged gradients, the relative log gradient magnitudes, and the relative log learning rate.

The full set of Parameter RNN outputs for each tensor $t$ are $\mb y^n_t  = \left\{ \mb d^n_{\theta t}, \mb d^n_{\phi t}, \Delta \eta^n_{\theta t}, \beta_{g t}^n, \beta_{\lambda t}^n \right\}$, corresponding to the parameter and attention update directions, the change in step length, and the momentum logits. Each of the outputs in $\mb y^n_t$ is read out via a learned affine transformation of the Parameter RNN hidden state. The readout biases are clamped to 0 for $\mb d^n_\theta$ and $\mb d^n_\phi$. The RNN update equations are then:
\begin{align}
    \mb h^{n+1}_\text{Param} &= \text{ParamRNN}(\mathbf{x}^n, \mathbf{h}^n_\text{Param}, \mathbf{h}^n_\text{Tensor}, \mathbf{h}^n_\text{Global}) \\
    \mb h^{n+1}_\text{Tensor} &= \text{TensorRNN}(\mathbf{x}^n, \mathbf{h}^{n+1}_\text{Param}, \mathbf{h}^n_\text{Tensor}, \mathbf{h}^n_\text{Global}) \\
    \mb h^{n+1}_\text{Global} &= \text{GlobalRNN}(\mathbf{x}^n, \mathbf{h}^{n+1}_\text{Param}, \mathbf{h}^{n+1}_\text{Tensor}, \mathbf{h}^n_\text{Global}) \\
    \mb y^{n} &= \mb W \mathbf{h}^n_\text{Param} + \mb b,
\end{align}
where $\mb h^n$ is the hidden state for each level of the RNN, as described in Section \ref{sec hier},
and $\mb W$ and $\mb b$ are learned weights of the affine transformation from the lowest level hidden state to output.


\subsection{Compute and memory cost}\label{sec compute cost}

The computational cost of the learned optimizer is $\mathcal{O}\left( N_P B + N_P K_P^2 + N_T K_T^2 + K_G^2 \right)$, where $B$ is the minibatch size, $N_P$ is the total number of parameters, $N_T$ is the number of parameter tensors, and $K_P$, $K_T$, and $K_G$ are the latent sizes for Parameter, Tensor, and Global RNNs respectively. Typically, we are in the regime where $N_P K_P^2 \gg N_T K_T^2 > K_G^2$, in which case the computational cost simplifies to $\mathcal{O}\left( N_P B + N_P K_P^2 \right)$. Note that as the minibatch size $B$ is increased, the computational cost of the learned optimizer approaches that of vanilla SGD, as the cost of computing the gradient dominates the cost of computing the parameter update.

The memory cost of the learned optimizer is $\mathcal{O}\left( N_P + N_P K_P + N_T K_T + K_G \right)$, which similarly to computational cost typically reduces to $\mathcal{O}\left( N_P + N_P K_P \right)$. So long as the latent size $K_P$ of the Parameter RNN can be kept small, the memory overhead will also remain small.


We show experimental results for computation time in Section \ref{sec wall clock}.

\section{Meta-training}
The RNN optimizer is meta-trained by a standard optimizer on an ensemble of target optimization tasks. We call this process meta-training, and the parameters of the RNN optimizer the meta-parameters.

\subsection{Meta-training set}\label{sec meta train set}

Previous learned optimizers have failed to generalize beyond the problem on which they were meta-trained. In order to address this, we meta-train the optimizer on an ensemble of small problems, which have been chosen to capture many commonly encountered properties of loss landscapes and stochastic gradients. By meta-training on small toy problems, we also avoid memory issues we would encounter by meta-training on very large, real-world problems. 

Except where otherwise indicated, all target problems were designed to have a global minimum of zero (in some cases a constant offset was added to make the minimum zero). 
The code defining each of these problems is included in the open source release. See \ref{open source}.



\subsubsection{Exemplar problems from literature}

We included a set of 2-dimensional problems which have appeared in optimization literature \cite{optimization-functions} as toy examples of various loss landscape pathologies. These consisted of Rosenbrock, Ackley, Beale, Booth, Styblinski-Tang, Matyas, Branin, Michalewicz, and log-sum-exp functions.

\subsubsection{Well behaved problems}

We included a number of well-behaved convex loss functions, consisting of quadratic bowls of varying dimension with randomly generated coupling matrices, and logistic regression on randomly generated, generally linearly separable data. For the logistic regression problem, when the data is not fully linearly separable, the global minimum is greater than 0.

\subsubsection{Noisy gradients and minibatch problems}

For problems with randomly generated data, such as logistic regression, we fed in minibatches of various sizes, from 10 to 200. We also used a minibatch quadratic task, where the minibatch loss consisted of the square inner product of the parameters with random input vectors.

For full-batch problems, we sometimes added normally distributed noise with standard deviations from 0.1 to 2.0 in order to simulate noisy minibatch loss. 

\subsubsection{Slow convergence problems}

We included several tasks where optimization could proceed only very slowly, despite the small problem size. This included a many-dimensional oscillating valley whose global minimum lies at infinity, and a problem with a loss consisting of a very strong coupling terms between parameters in a sequence. 
We additionally included a task where the loss only depends on the minimum and maximum valued parameter, so that gradients are extremely sparse and the loss has discontinuous gradients.


\subsubsection{Transformed problems}

We also included a set of problems which transform the previously defined target problems in ways which map to common situations in optimization.

To simulate problems with sparse gradients, one transformation sets a large fraction of the gradient entries to 0 at each training step. 
To simulate problems with different scaling across parameters, we added a transformation which performs a linear change of variables so as to change the relative scale of parameters. 
To simulate problems with different steepness-profiles over the course of learning, we added a transformation which applied monotonic transformations (such as raising to a power) to the final loss.
Finally, to simulate complex tasks with diverse parts, we added a multi-task transformation, which summed the loss and concatenated the parameters from a diverse set of problems.



\subsection{Meta-objective}\label{sec meta-obj}

For the meta-training loss, used to train the meta-parameters of the optimizer, we used the average log loss across all training problems,
\begin{align} \label{eq metaloss}
L\left(\psi \right) = \frac{1}{N}\sum_{n=1}^{N} \left(
\log \left( \ell(\theta^n\left(\psi\right) ) + \epsilon \right) 
- \log \left( \ell(\theta^0) + \epsilon \right) 
\right)
,
\end{align}
where the second term is a constant, and where $\psi$ is the full set of meta-parameters for the learned optimizer, consisting of $\psi = \left\{ \psi_\text{P-RNN}, \psi_\text{T-RNN}, \psi_\text{G-RNN}, \gamma, c \right\}$, where $\psi_{\bullet\text{-RNN}}$ indicates the GRU weights and biases for the Parameter, Tensor, or Global RNN, $\gamma$ is the learning rate momentum and $c$ is the attended step offset (Section \ref{sec decomposition}).

Minimizing the average log function value, rather than the average function value, better encourages exact convergence to minima and precise dynamic adjustment of learning rate based on gradient history (Figure \ref{fig ablation}). The average logarithm also more closely resembles minimizing the {\em final} function value, while still providing a meta-learning signal at every training step, since very small values of $\ell(\theta^n)$ make an outsized contribution to the average after taking the logarithm.




\subsection{Partial unrolling}

Meta-learning gradients were computed via backpropagation through partial unrolling of optimization of the target problem, similarly to \citet{andrychowicz2016learning}. Note that \citet{andrychowicz2016learning} dropped second derivative terms from their backpropagation, due to limitations of Torch. We compute the full gradient in TensorFlow, including second derivatives. 

\subsection{Heavy-tailed distribution over training steps} 
\label{sec num train steps}

In order to encourage the learned optimizer to generalize to long training runs, both the number of partial unrollings, and the number of optimization steps within each partial unroll, was drawn from a heavy tailed exponential distribution. The resulting distribution is shown in Appendix \ref{sec appendix_num_train_steps}



\subsection{Meta-optimization}\label{sec meta opt}

The optimizers were meta-trained for at least 40M meta-iterations (each meta-iteration consists of loading a random problem from the meta-training set, running the learned optimizer on that target problem, computing the meta-gradient, and then updating the meta-parameters). 
The meta-objective was minimized with asynchronous RMSProp across 1000 workers, with a learning rate of $10^{-6}$.

\section{Experiments}
\begin{figure}[t]
\includegraphics[width=\columnwidth]{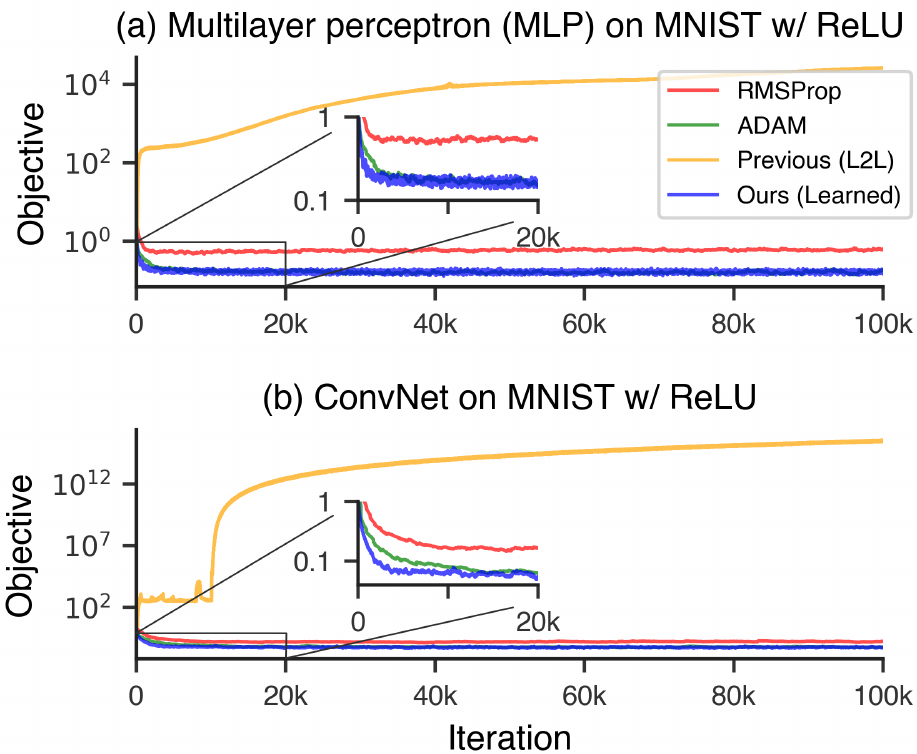}
\caption{Training loss versus number of optimization steps on MNIST for the Learned optimizer in this paper compared to the L2L optimizer from \citet{andrychowicz2016learning}, ADAM (learning rate 2e-3), and RMSProp (learning rate 1e-2). The L2L optimizer from previous work was meta-trained on a 2-layer, fully-connected network with sigmoidal nonlinearities. The test problems were a 2-layer fully-connected network and a 2-layer convolutional network. In both cases, ReLU activations and minibatches of size 64 was used.}
\label{fig opt diverge}
\end{figure}
\begin{figure}[t]
\includegraphics[width=\columnwidth]{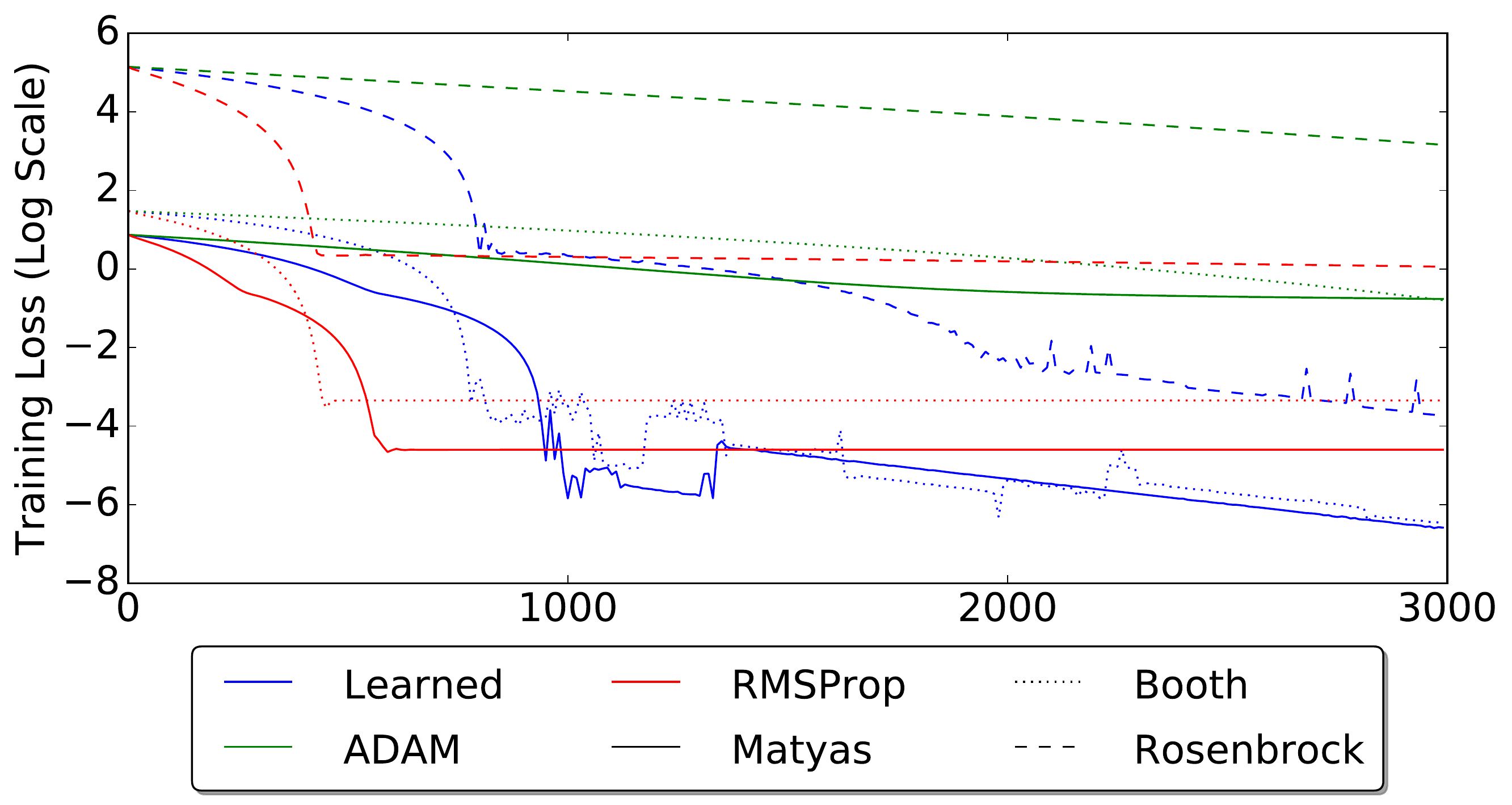}
\caption{Three sample problems from the meta-training corpus on which the learned optimizer outperforms RMSProp and ADAM. The learning rates for RMSProp (1e-2) and ADAM (2e-3) were chosen for good average performance across all problem types in the training and test set. The learned optimizer generally beats the other optimizers on problems in the training set.}
\label{fig train set perf}
\end{figure}
\begin{figure*}[t]
\begin{subfigure}[h]{\columnwidth}
    \includegraphics[width=\columnwidth]{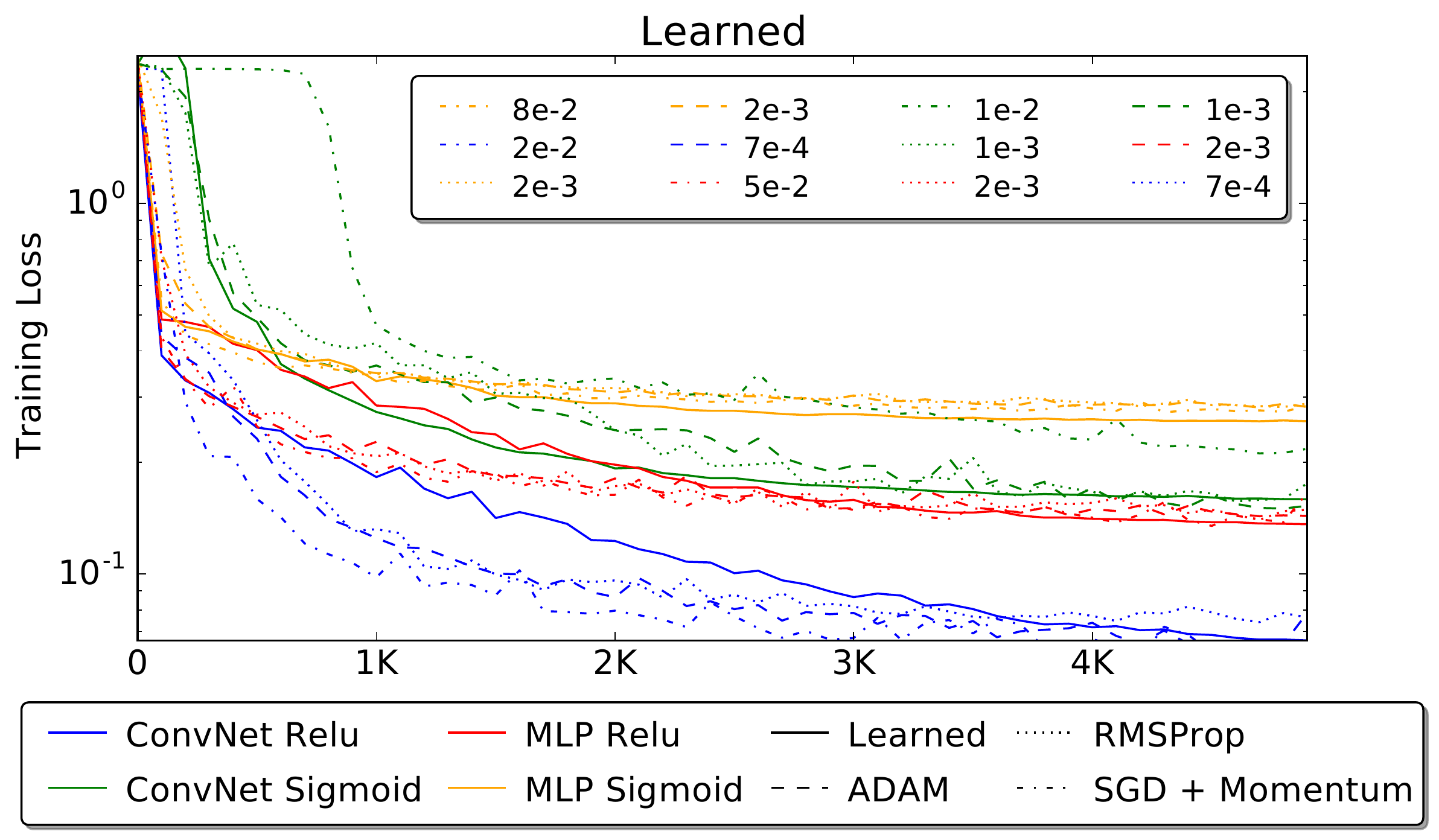}
    \caption{Learned optimizer matches performance of ADAM, RMSProp, and SGD with momentum on four problems never seen in the meta-training set. For the non-learned optimizer, the optimal learning rate for each problem was chosen from a sweep over learning rates from $10^{-9}$ to $0.1$. Actual learning rates used are shown in the inset legend.}
    \label{fig test set perf}
\end{subfigure}
\hfill
\begin{subfigure}[ht]{\columnwidth}
    \includegraphics[width=\columnwidth]{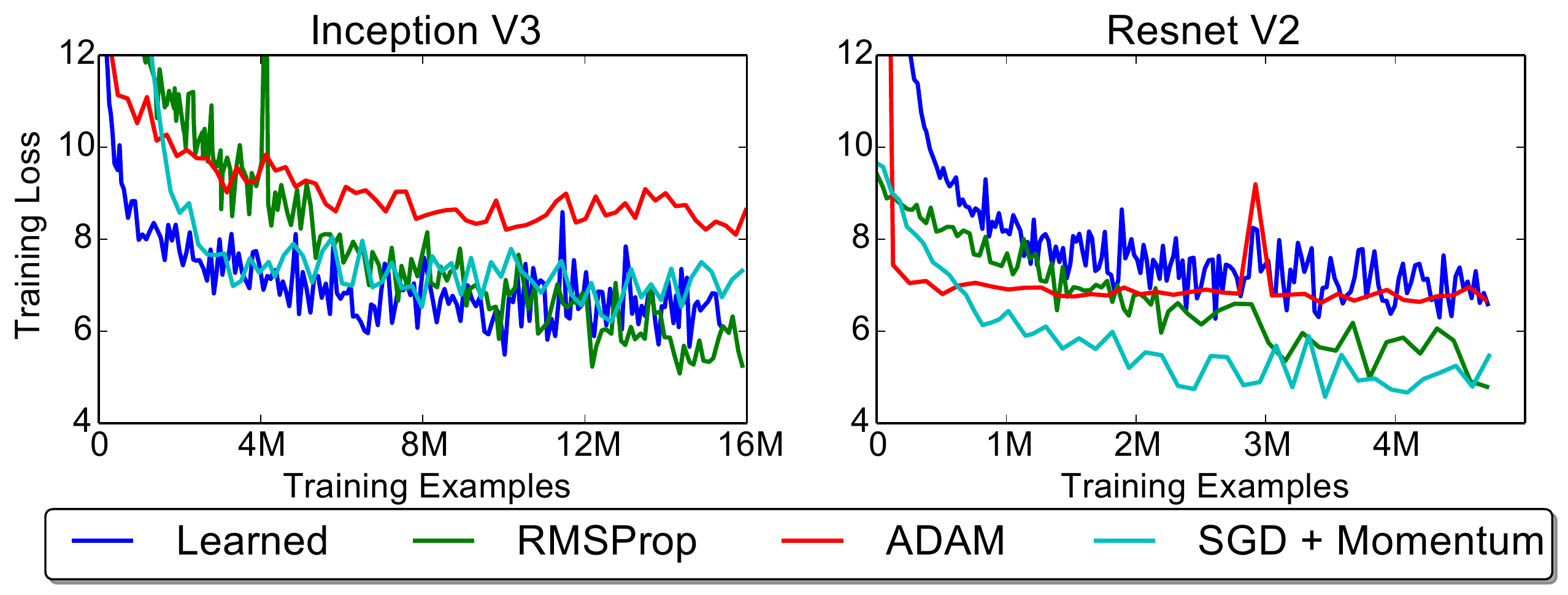}
    \caption{Training loss on ImageNet data in early training as a function of number of training examples seen (accounting for varying minibatch sizes). While other optimizer performance is highly dependent on hyperparameters, learned optimizer performance is similar to the best tuned optimizers (though in late training, the learned optimizer loss increases again). In both cases the learned optimizer was used for distributed, synchronized learning with an effective minibatch size of 800. The Inception V3 plot was generated from a newer version of the codebase, with small improvements described in Appendix \ref{app lol future}. On Inception V3, other optimizers used a learning rate of 0.045 and an effective minibatch size of 1600 (the optimal hyperparameters for the RMSProp optimizer from the original paper). On Resnet, other optimizers used a learning rate of 0.1 and an effective minibatch size of 256 (the optimal hyperparameters for the SGD + momentum optimizer from the original paper).}
    \label{fig inception}
\end{subfigure}
\caption{The learned optimizer generalizes to new problem types unlike any in the meta-training set, and with many more parameters.}
\end{figure*}
\subsection{Failures of existing learned optimizers}

Previous learned optimizer architectures like \citet{andrychowicz2016learning} perform well on the problems on which they are meta-trained. However, they do not generalize well to new architectures or scale well to longer timescales. Figure~\ref{fig opt diverge} shows the performance of an optimizer meta-trained on a 2-layer perceptron with sigmoid activations on the same problem type with ReLU activations and a new problem type (a 2-layer convolutional network). In both cases, the same dataset (MNIST) and minibatch size (64) was used. In contrast, our optimizer, which has not been meta-trained on this dataset or {\em any} neural network problems,
shows performance comparable with ADAM and RMSProp, even for numbers of iterations not seen during meta-training (Section~\ref{sec num train steps}).

\subsection{Performance on training set problems}
The learned optimizer matches or outperforms ADAM and RMSProp on problem types from the meta-training set (Figure~\ref{fig train set perf}). The exact setup for each problem type can be seen in the python code in the supplementary materials.


\subsection{Generalization to new problem types}

The meta-training problem set did not include any convolutional or fully-connected layers. Despite this, we see comparable performance to ADAM, RMSProp, and SGD with momentum on simple convolutional multi-layer networks and multi-layer fully connected networks both in terms of final loss and number of iterations to convergence (Figure \ref{fig test set perf} and Figure \ref{fig opt diverge}).

We also tested the learned optimizer on Inception V3 \cite{szegedy2016rethinking} and on ResNet V2 \cite{he2016identity}. 
Figure \ref{fig inception} shows the learned optimizer is able to stably train these networks for the first 10K to 20K steps, with performance similar to traditional optimizers tuned for the specific problem. 
Unfortunately, we find that later in training the learned optimizer stops making effective progress, and the loss approaches a constant (approximately 6.5 for Inception V3). 
Addressing this issue would be a goal of future work.


\subsection{Performance is robust to choice of learning rate}
\begin{figure}[h!]
\includegraphics[width=\columnwidth]{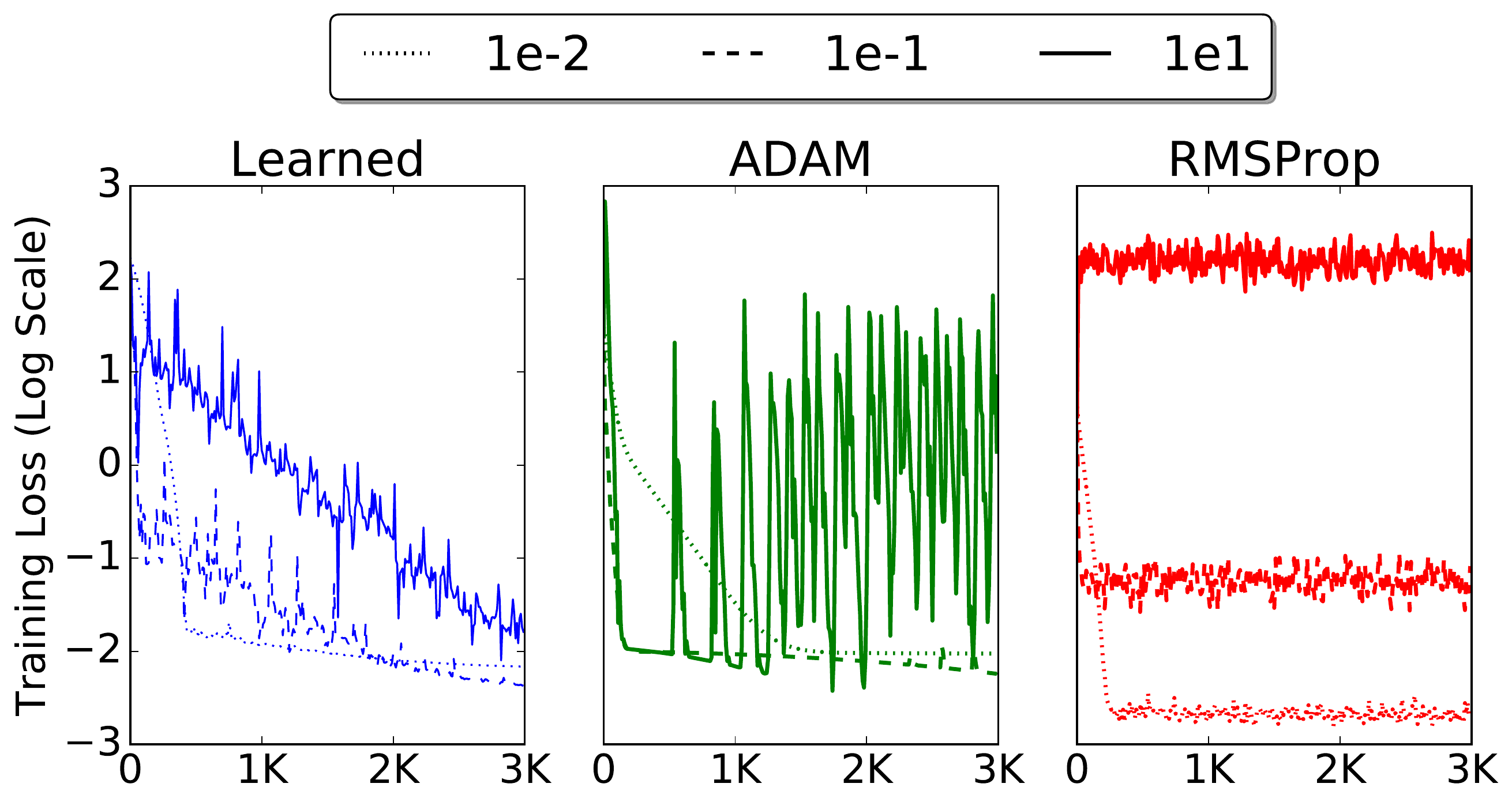}
\caption{Learned optimizer performance is robust to learning rate hyperparameter. Training curves on a randomly generated quadratic loss problem with different learning rate initializations.}\label{fig lr init}
\end{figure}
One time-consuming aspect of training neural networks with current optimizers is choosing the right learning rate for the problem. While the learned optimizer is also sensitive to initial learning rate, it is much more robust. Figure \ref{fig lr init} shows the learned optimizer's training loss curve on a quadratic problem with different initial learning rates compared to those same learning rates on other optimizers. 

\subsection{Ablation experiments} \label{sec ablation}
\begin{figure}[ht]
\includegraphics[width=\columnwidth]{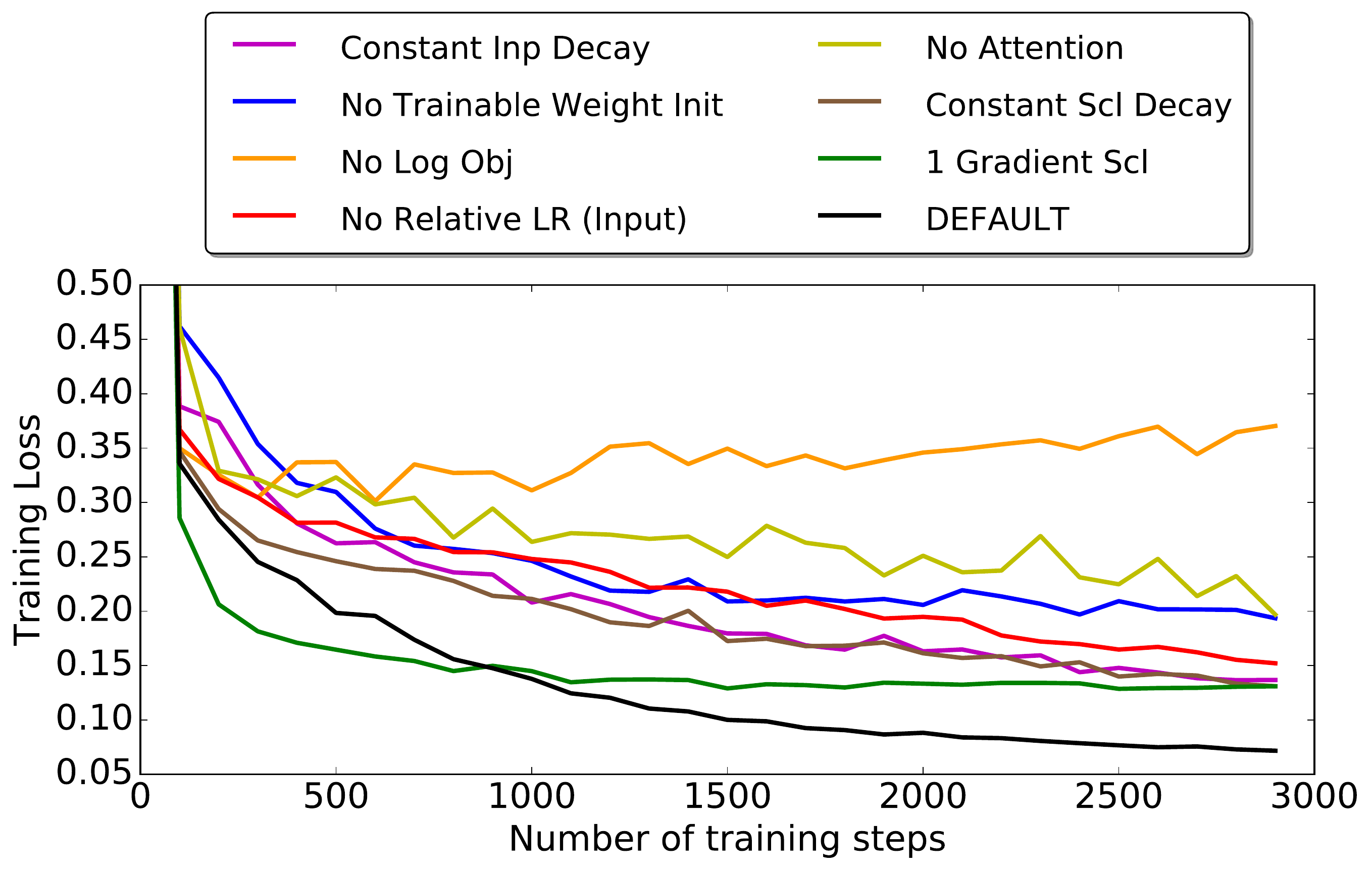}
\caption{Ablation study demonstrating importance of design choices on a small ConvNet on MNIST data. DEFAULT is the optimizer with all features included.}
\label{fig ablation}
\end{figure}
The design choices described in Section \ref{sec arch} matter for the performance of the optimizer. We ran experiments in which we removed different features and re-meta-trained the optimizer from scratch. We kept the features which, on average, made performance better on a variety of test problems. Specifically, we kept all of the features described in \ref{sec features} such as attention (\ref{sec attention}), momentum on multiple timescales (gradient scl) (\ref{sec multi_timescale}), dynamic input scaling (variable scl decay) (\ref{sec dyn inp scl}), and a relative learning rate (relative lr) (\ref{sec decomposition}). We found it was important to take the logarithm of the meta-objective (log obj) as described in \ref{sec meta-obj}. In addition, we found it helpful to let the RNN learn its own initial weights (trainable weight init) and an accumulation decay for multiple gradient timescales (inp decay).
Though all features had an effect, some features were more crucial than others in terms of consistently improved performance. Figure \ref{fig ablation} shows one test problem (a 2-layer convolutional network) on which all final features of the learned optimizer matter. 

\subsection{Wall clock comparison} \label{sec wall clock}

In experiments, for small minibatches, we significantly underperform ADAM and RMSProp in terms of wall clock time. However, consistent with the prediction in \ref{sec compute cost}, since our overhead is constant in terms of minibatch we see that the overhead can be made small by increasing the minibatch size.

\begin{figure}[ht]
\includegraphics[width=\columnwidth]{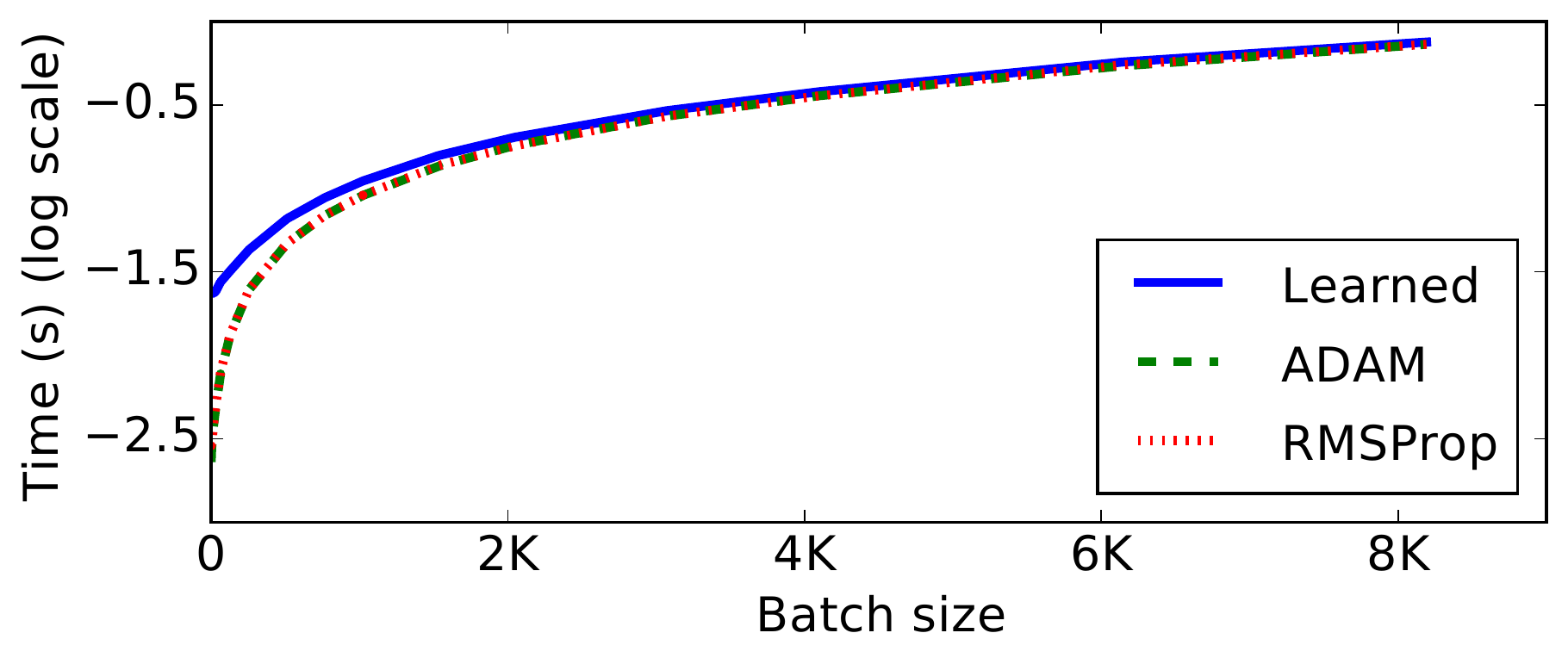}
\caption{Wall clock time in seconds to run a single gradient and update step for a 6-layer ConvNet architecture on an HPz440 workstation with an NVIDIA Titan X GPU. As batch size increases, the total computation time for the Learned optimizer approaches ADAM.}

\label{fig wall clock}
\end{figure}

\section{Conclusion}
We have shown that RNN-based optimizers meta-trained on small problems can scale and generalize to early training on large problems like ResNet and Inception on the ImageNet dataset. To achieve these results, we introduced a novel hierarchical architecture that reduces memory overhead and allows communication across parameters, and augmented it with additional features shown to be useful in previous optimization and recurrent neural network literature. We also developed an ensemble of small optimization problems that capture common and diverse properties of loss landscapes. Although the wall clock time for optimizing new problems lags behind simpler optimizers, we see the difference decrease with increasing batch size. Having shown the ability of RNN-based optimizers to generalize to new problems, we look forward to future work on optimizing the optimizers.

\newpage
\clearpage

\bibliography{LOL_ICML_2017}
\bibliographystyle{icml2017}

\newpage
\clearpage

\appendix
\part*{Appendix}

\setcounter{figure}{0} \renewcommand{\thefigure}{App.\arabic{figure}}
\setcounter{table}{0} \renewcommand{\thetable}{App.\arabic{table}}

\section{Code}\label{open source}
The code for the meta-training procedure and meta-train problem set is available at \href{url}{https://git.io/v5oq5}. 

\section{Additional details of RNN architecture}

\subsection{Shortcut connection}

\label{sec shortcut}

Since we expect $\mb m^n_{ts}$ to be the primary driver of update step direction, and in order to further reduce the information which must be stored in the Parameter RNN hidden state, we included a meta-trainable linear projection from the average rescaled gradients $\mb m^n_{ts}$ and the update directions $\Delta \mb \theta^n_t$ and $\Delta \mb \phi^n_t$.




\section{Additional details of meta-training process}

\subsection{Heavy-tailed distribution over training steps} 
\label{sec appendix_num_train_steps}

\begin{figure}[hhh]
\centering
\includegraphics[width=\columnwidth]{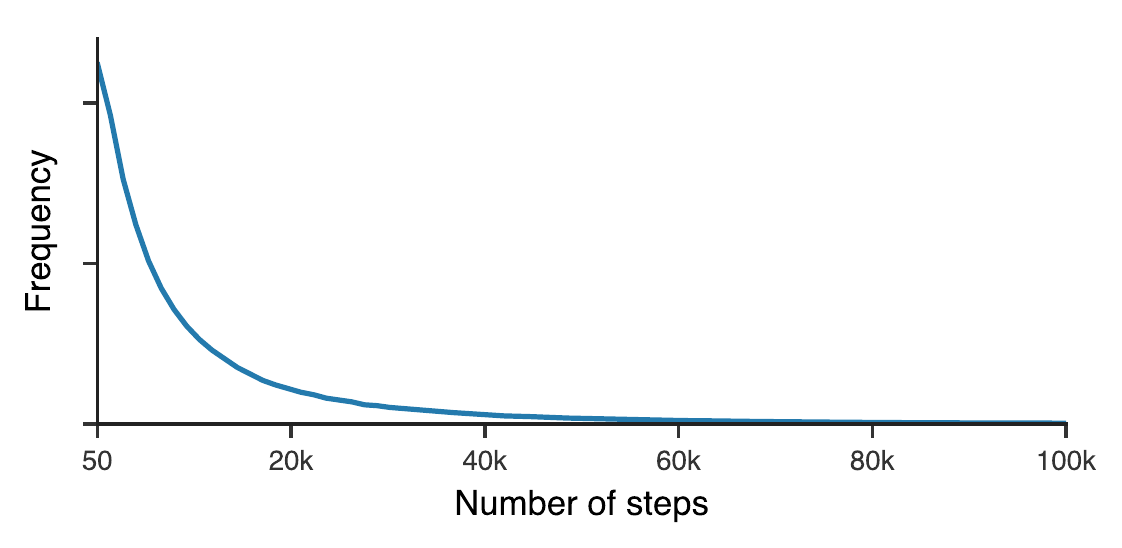}
\caption{A histogram of the total number of training iterations run on target problems during meta-training. The total number of unrolls is drawn from an exponential distribution with scale 50 plus a constant offset of 1. The number of training iterations within each unroll is drawn from an exponential distribution with scale 200 and a constant offset of 50. 
}
\label{fig num_iters}
\end{figure}

\section{Architecture updates}
\label{app lol future}

The Inception V3 experiment in Figure \ref{fig inception} used a slightly newer version of the learned optimizer codebase. The changes were:

\subsection{Parameter noise during training}

Due to the use of small meta-training problems in Section \ref{sec meta train set}, during meta-training the learned optimizer is often able to optimize the problem almost exactly early in the unrolled optimization, after which the meta-loss s becomes relatively uninformative. In order to better simulate tasks which take many steps to optimize, small Gaussian noise is added to the parameters during each optimization step. This effectively moves the loss landscape underneath the optimizer, providing a more informative learning signal after many unrolls, and forcing the learned optimizer to be robust to a new type of noise. Specifically, the parameter update becomes
\begin{align}
\theta^{n+1}_t &= \theta^n_t + \Delta \theta_t^n + \alpha \tilde{\mb n}^t \\
\tilde{\mb n} &\sim \mc N\left(\mb 0, \mb I \right)
\end{align}
where the noise scale $\alpha$ is drawn from a log uniform distribution between $10^{-10}$ and $10^{-2}$ for each problem.

\subsection{Momentum from {\em previous} timescale}

In Equation \ref{eq grads scaled} we scale the average gradients $\bar{\mb g}_{ts}^n$ by a running estimate $\sqrt{\lambda^{n}_{ts}}$ of the root-mean-square magnitude of $\bar{\mb g}_{ts}^n$. This is a mismatch with Adam, where the average gradient is scaled by a running estimate of the root-mean-square magnitude of the {\em non-averaged} gradients. 
In order to be consistent with this, and in order to encourage better use of the dynamic range of $\mb m^n_{ts}$ (as defined in the text body, it spends much of its time with values near $1$ or $-1$), we modify Equation \ref{eq grads scaled} to normalize the average gradient $\bar{\mb g}_{ts}^n$ by $\sqrt{\lambda^{n}_{ts}}$ from the immediately faster timescale,
\begin{align}
    \mb m^n_{ts} &= \frac{ \bar{\mb g}_{ts}^n }{ \sqrt{\lambda^{n}_{t(s-1)}} },
\end{align}
and where we define the average gradient at the fastest time scale to be the raw gradient, $\bar{\mb g}_{t(-1)}^n = \mb g^n_t$

\subsection{No normalization of step length}

In order to simplify interactions between parameters, we no longer force a normalization of the parameter and attention update directions $\mb d^n_{\theta t}$ and $\mb d^n_{\phi t}$. We {\em do} still decompose the update into the product of a learning rate and a step. Since the attended update direction is now able to take on a different magnitude, the separate attention log learning rate $\eta^n_\phi$ is no longer required, and is eliminated. Equations \ref{eq update step} and \ref{eq attend step} thus become
\begin{align}
    \Delta \mb \theta^n_t &= \exp\left(\eta^n_{\theta t} \right) 
        \mb d^n_{\theta t}
        , \\
    \Delta \mb \phi^n_t &= \exp\left(\eta^n_{\theta t} \right) 
        \mb d^n_{\phi t}
        .
\end{align}

\subsection{More stable meta-training hyper-parameters}

The distribution over meta-loss gradients is observed to be assymmetrical and heavy tailed. 
This combination is known to cause biased parameter updates in RMSProp and Adam, since both optimizers underweight the contribution from extremely rare extremely large gradients. 
In order to reduce this tendency, we updated the mean-quare-gradient momentum term $\gamma$ to be 0.999, rather than 0.9 in the meta-optimizer RMSProp (Section \ref{sec meta opt}).

\end{document}